# Recommender System Based on Algorithm of Bicluster Analysis RecBi

Dmitry Ignatov[2], Jonas Poelmans[1], Vasily Zaharchuk[1,2]

[1]K.U.Leuven, Faculty of Business and Economics, Naamsestraat 69,
3000 Leuven, Belgium
[5]National Research University Higher School of Economics (HSE), Pokrovskiy boulvard 11
101000 Moscow, Russia
Dignatov@hse.ru
Jonas.Poelmans@econ.kuleuven.be

**Abstract.** In this paper we propose two new algorithms based on biclustering analysis, which can be used at the basis of a recommender system for educational orientation of Russian School graduates. The first algorithm was designed to help students make a choice between different university faculties when some of their preferences are known. The second algorithm was developed for the special situation when nothing is known about their preferences. The final version of this recommender system will be used by Higher School of Economics.

## 1 Introduction

Since the introduction of the so called Common State Exam in high schools of the Russian Federation, graduates received permission to apply to enter multiple universities or faculties of the same university whereas in the past they were only allowed to apply to one institution. Students are confronted with an ever increasing complexity of the educational landscape and for this purpose we developed a recommender system to guide them in their search. Students can indicate one or more faculties where they would like to study and our recommender system will make suggestions on alternative institutions in which they might also be interested. The recommender system will also use the browsing and searching history of the candidate student to efficiently suggest relevant universities, faculties, and educational directions.

Currently on the Internet many websites make use of recommender systems, for example, Amazon recommends books in which the client might be interested based on previous items which were viewed by the user. Other examples include the websites http://facebook.com/ and http://twitter.com/ and for Russian companies, the websites http://imhonet.ru/ and http://www.ozon.ru/.

A lot of techniques have been developed for recommender systems and the main principles of these algorithms are described in [5]. We can distinguish between item-based and user-based recommender systems. In item-based recommendation relevant items are presented based on their similarity to items previously accessed or bought by the user. In user-based recommendation users with a similar profile to the current user are gathered and based on the items they accessed or bought relevant suggestions are made. These systems use different kinds of similarity measures, such as Pearson's correlation, Euclidean distance, Jacquard coefficient, and Manhattan distance.

One of the most recent innovations in recommender system research is applying methods based on biclustering. In [1-5] a wide range of biclustering applications has been described including market research, near-duplicate web-document detection, bioinformatics etc. Biclustering is an unsupervised learning method similar to Formal Concept Analysis (FCA) [7, 8, 9].

Comparing to traditional clustering methods biclustering is not a blackbox technique. Comprehensibility is one of its main advantages, i.e. it is possible to understand why objects ended up in the same cluster. For example you might ask why a cucumber and a pair of boots are assigned to the same cluster. With biclustering it can easily be revealed that they are similar because they have the same color and skin surface.

This lack of comprehensibility of traditional clustering techniques may cause serious problems in large data mining projects. To cope with these issues researchers are increasingly focusing on human-centered techniques including direct clustering (John Hartigan's work [6]) and FCA [10]. In this paper we chose to use biclustering instead of the more famous technique FCA because of the scalability issues encountered with FCA. The large number of extracted concepts quickly results in an unreadable lattice.

## 2 Algorithms

We will use the general biclustering definition, which was given in [1, 3]. Let $A$ be a matrix of size ($n \times m$), where $m$ represents the dimensionality of the set of objects and $n$ represents the dimensionality of the set of attributes. Then $X = \{x_1, x_2, ..., x_n\}$ is a set of objects and $Y = \{y_1, y_2, ..., y_m\}$ is a set of attributes. If $I \subseteq X$ and $J \subseteq Y$, then $A_{IJ}$ is a submatrix of matrix $A$. $A_{IY} = (I, Y)$ is a cluster of objects of matrix $A$ and $A_{XJ} = (X, J)$ is a cluster of attributes of matrix $A$. $A_{IJ} = (I, J)$ is a bicluster of matrix $A$. Its objects share similar attributes and its attributes give a description of the objects in the cluster.

There are different formal definitions of a bicluster available for several specific cases, however these are not considered in this paper.

### 2.1 Algorithm variant RecBi1

RecBi1 is based on the algorithm of Ignatov D.I., which was described in [4]. This algorithm takes as input two contexts and produces a list of recommendations as output.

Context 1: Formal context $K = (S, A, I)$, where $S$ is a list of all faculties of Russian universities, $A$ is a list of faculty attributes, $I$ is a binary relation, that shows that faculty $s$ from $S$ has an attribute $a$ from $A$.

Context 2: Multi-valued context containing the history of the usage of the system $K_w = (U, A, W, J)$, where $U$ is a list of users, $A$ is a list of faculty attributes, $W$ is a list of weights, that shows how many times $u$ from $U$ has looked at and considered $a$ from $A$ as an interesting item, $J$ is a ternary relation between $u$, $a$, and $w$.

| **RecBi1** |
|---|
| **Input:** Formal context of faculties $K = (S, A, I)$, <br> Multi-valued context containing history of usage $K_w = (U, A, W, J)$, <br> Visits vector $V=(V_1,\ldots,V_{|U|})$, <br> $U_0$ is a target user, $N$ is a number of recommendations. |
| **Output:** Rec is a list of recommendations |
| 1. $(U_0, s_i)$ // initial couple <br> 2. For $s_i$ in $S \neq \emptyset$ // $s_i'$ are attributes of $s_i$ <br> $\quad$ CandidateS($cs$) $\cup$ $(s_i, s_j' \cap s_i')$ <br> 3. For $cs_i$ in $cs$ <br> $$kc[cs] = \frac{\sum_{attr \in cs_i[1]} \frac{attr(u_0)}{v_{u_0}}}{|cs_i'|}$$ <br> 4. $kc = \text{dec\_sort}(kc)$ <br> 5. $Rec = \text{Top}(N, kc)$ <br> Return $Rec$ |

## 2.2 Algorithm variant RecBi2

The second algorithm variant consists of two parts: Recbi2.1 and RecBi2.2. RecBi2.1 is used with so called cold start, which means that there is no previous usage history available. RecBi2.2 is used when the user is using this system not for the first time.

RecBi2.1 takes the same contexts as input as RecBi1 and outputs a list of recommendations. But in the middle of the algorithm it forms a new formal context.

Context 3: Formal context of user preferences $K_{lp} = (U, S, Z)$, where $U$ is a list of users, $S$ is a list of faculties, and $Z$ is a binary preference relation between $u$ from $U$ and $s$ from $S$.

| **RecBi2.1** |
|---|
| **Input:** Formal context of faculties $K = (S, A, I)$, <br> Multi-valued context containing history of usage $K_w = (U, A, W, J)$, <br> Visits vector $V=(V_1,\ldots,V_{|U|})$, <br> $U_0$ is a target user, $N$ is a number of recommended items, $l_{min}$ is a minimal index of |

| |
|---|
| interestingness. |
| **Output:** *Rec* is a non-ranked list |
| 1. $K_1 = K_w \cdot K^T = (U, S, L, Y)$ //multiplication of multi-valued contexts as a matrix<br>2. $Z = \emptyset$ ; $K_{pl} = (U, S, Z \subseteq U \times S)$<br>3. For *u* in *U*<br>    For *s* in *S*<br>        If $scu \geq l_{min}$ then $Z=(s, u) \cup Z$ //reduction to formal context<br>4. $(U_0, s)$ //*s'* is a set of all visitors of *s*<br>5. For *u* in *s'*<br>    $S_c = \bigcup u'$<br>6. For $s_c$ in $S_c$<br>    $V(s_c) = \|s_c' \cap S\|$<br>7. $v = \text{dec\_sort}(v)$<br>8. $Rec = \text{Top}(N, v)$<br>Return *Rec* |

RecBi2.2 works in a similar way as RecBi2.1, but its output is a ranked list of search results based on user feedback.

## 3 Conclusion

In this paper we presented two biclustering-based algorithms which will be used at the basis of recommender system which will guide high school students in their search for an educational institution. The main contribution of this paper is a successful application of biclustering methods to the educational domain. In the future we will operationalise this system at the Higher School of Economics.